\documentclass[pdflatex,sn-apa]{sn-jnl}

\usepackage[utf8]{inputenc}
\usepackage{graphicx}%
\usepackage{multirow}%
\usepackage{amsmath,amssymb,amsfonts}%
\usepackage{amsthm}%
\usepackage{mathrsfs}%
\usepackage[title]{appendix}%
\usepackage{xcolor}%
\usepackage{textcomp}%
\usepackage{manyfoot}%
\usepackage{array}%
\usepackage{booktabs}%
\usepackage{anyfontsize}%
\usepackage{tabularx}
\usepackage{url}
\usepackage{xspace}
\newcommand{\ie}{i.e.,\xspace}

\theoremstyle{thmstyleone}%

\theoremstyle{thmstyletwo}%

\theoremstyle{thmstylethree}%
\newtheorem{definition}{Definition}%

\title[Article Title]{A Taxonomy on Collective Awareness}

\author*[1,2]{\fnm{Guillermo} \sur{GP-Lenza} \orcid{https://orcid.org/0009-0004-6245-8380}}\email{g.glenza@upm.es}

\author[2,4]{\fnm{Miguel} \sur{Fernandez-Cortizas} \orcid{https://orcid.org/0000-0002-3822-075X}}\email{miguel.fernandez.cortizas@upm.es}

\author[1,2]{\fnm{Martin} \sur{Molina} \orcid{https://orcid.org/0000-0001-7145-1974}}\email{martin.molina@upm.es}

\author[3]{\fnm{Ricardo} \sur{Sanz} \orcid{https://orcid.org/0000-0002-2381-933X}} \email{ricardo.sanz@upm.es}

\author[2]{\fnm{Pascual} \sur{Campoy} \orcid{https://orcid.org/0000-0002-9894-2009}} \email{pascual.campoy@upm.es}

\affil[1]{\orgdiv{Computer Vision and Aerial Robotics Group}, \orgname{Department of Artificial Intelligence, Universidad Politécnica de Madrid}, \orgaddress{\street{C. de los Ciruelos}, \city{Boadilla del Monte}, \postcode{28660}, \state{Madrid}, \country{Spain}}}

\affil[2]{\orgdiv{Computer Vision and Aerial Robotics Group}, \orgname{Centre for Automation and Robotics (CAR), Universidad Politécnica de Madrid (UPM-CSIC)}, \orgaddress{\street{Calle José Gutiérrez Abascal 2}, \city{Madrid}, \postcode{28006}, \state{Madrid}, \country{Spain}}}

\affil[3]{\orgdiv{Autonomous Systems Laboratory}, \orgname{ASLAB, Universidad Politécnica de Madrid (UPM-CSIC)}, \orgaddress{\street{Calle José Gutiérrez Abascal 2}, \city{Madrid}, \postcode{28006}, \state{Madrid}, \country{Spain}}}

\affil[4]{\orgdiv{Automation and Robotics Research Group (ARG)}, \orgname{Interdisciplinary Centre for Security, Reliability, and Trust (SnT), University of Luxembourg}, \orgaddress{\postcode{L-1359}, \city{Luxembourg}, \country{Luxembourg}}}

\abstract{As robotic teams tackle increasingly complex tasks in dynamic and unstructured environments, effective coordination requires agents to maintain accurate, aligned representations of their environment, teammates, and mission state. We argue that Mutual Awareness, Shared Situational Awareness, and Team Situational Awareness --- three concepts widely invoked in the multi-robot systems literature --- are not interchangeable: they form a containment hierarchy in which each type subsumes the previous in scope, and span a distribution spectrum from fully individualized understanding (MA) to fully uniform understanding (SSA), with TSA occupying a mixed position. We establish this through a two-dimensional taxonomy organized along the object of awareness and awareness distribution axes. Treating these terms as synonyms obscures the precise coordination requirements each imposes on a robotic system. A Search and Rescue case study with heterogeneous aerial and ground robots grounds each taxonomy position in concrete coordination requirements. These findings provide a conceptual foundation for principled specification, design, and comparison of collective awareness in multi-robot systems.}


\begin{document}
\maketitle

\section{Introduction}

Multi-agent systems in robotics represent an innovative approach to solving complex problems through the coordination and collaboration of multiple robotic units exhibiting emergent behaviors. These systems offer several advantages, including increased robustness, scalability, and flexibility in tackling diverse challenges across various domains, such as area mapping, collective construction, and target search and tracking. By leveraging collaboration and cooperation between its members, these robotic systems can achieve tasks that would be difficult or impossible for individual robots to accomplish alone, making them particularly useful in many applications ranging from automated construction to environmental monitoring~\cite{WoolridgeIntroMAS}.

More formally, a Multi-Agent System (MAS) is a collection of agents that concurrently sense and act upon an environment, transforming it through their combined activities, with emergent behavior arising from agent interactions rather than from any individual agent in isolation~\cite{WeissMAS}. A multi-robot system (MRS) constitutes a specific instantiation of this paradigm in which all agents are embodied robots interacting within a shared physical environment to accomplish tasks that exceed the capabilities of any individual unit~\cite{MRSCoordination}. MRS vary along several dimensions: they may be homogeneous or heterogeneous in robot composition, communicating or non-communicating in information exchange, and cooperative or competitive in their interaction objectives~\cite{MRSClassification}. Enabling robotic teams to pursue shared objectives without unnecessary overhead, such as resolving conflicts, redundantly synchronizing actions, or correcting misalignments, requires coordination~\cite{WoolridgeIntroMAS}~\cite{BondGasser}. Awareness plays a central role in enabling this coordination, as agents must maintain accurate representations of their environment, their teammates, and the evolving mission state to anticipate and complement each other's actions.

A key element of agent autonomy is the theoretical foundation laid by~\cite{Endsley1995} about Situation Awareness (SA) in the form of a model divided into three layers: perception of the environmental elements, comprehension of their meaning, and projection of their future status. In recent theoretical work, awareness has been proposed as the real-time understanding of sensory flows, where understanding is achieved when an agent successfully derives valid inferences about objects in its environment~\cite{CSAwareness}. This framework enables individual agents to make informed decisions and act appropriately within their environment. Building on this foundation, concepts such as Mutual Awareness, Shared Situational Awareness, and Team Situational Awareness emerged to characterize how team members align their understanding to enable effective coordination in multi-agent contexts. 

These concepts have been widely adopted by the multi-robot systems community, where researchers frequently invoke awareness as a fundamental notion in coordination capabilities~\cite{FernandezCortizas2024Landscape}. However, the three terms are often used interchangeably, without distinguishing the distinct requirements each imposes on a robotic system. We argue this conflation is consequential: MA, SSA, and TSA differ both in the scope of what agents must know and in whether that knowledge must be uniform across the team. Clarifying these distinctions raises two questions that remain underspecified in the literature:

\begin{enumerate}
    \item What are the distinct requirements of the different awareness types, and how do they relate to each other?
    \item What precise definitions can be established for each awareness type that characterize their distinct requirements and roles in multi-robot coordination?
\end{enumerate}

Addressing these questions is essential for enabling principled system design and technical progress in multi-robot coordination. Drawing from cognitive science, multi-agent systems, and robotics, this article examines these three awareness types and develops a conceptual framework to clarify their distinct requirements and relationships. Section~\ref{sectionRevAwareness} reviews the three main forms of awareness identified in the literature, Mutual Awareness, Shared Situational Awareness, and Team Situational Awareness, analyzing their definitions and key distinctions. Section~\ref{section:CATaxonomy} presents a taxonomy that characterizes these awareness types along two axes, awareness distribution and object of awareness, and illustrates each classification using a Search and Rescue scenario involving a heterogeneous team of aerial and ground robots. Finally, Section~\ref{section:Conclusions} synthesizes the main findings and identifies directions for future research.

\section{Awareness in Multi-Robot Systems} \label{sectionRevAwareness}

The notion of Situation Awareness (SA), originally formulated by Endsley~\cite{Endsley1995} as a three-layer model encompassing the perception of environmental elements, the comprehension of their meaning, and the projection of their future status, has been progressively extended to the domain of multi-robot systems. In this collective context, individual SA provides the foundation upon which distributed forms of awareness are built, as each agent must maintain its own understanding of the environment and its role within the team. Researchers across cognitive science, human factors, and multi-agent systems have proposed a range of overlapping concepts to characterize how awareness emerges and is maintained across a robotic collective. This section reviews the three forms that have received the widest attention in the multi-robot systems literature: Mutual Awareness, Shared Situational Awareness, and Team Situational Awareness.\footnote{Both ``Situation Awareness'' (Endsley's original formulation) and ``Situational Awareness'' (as used in the names of its collective extensions) appear in the literature; this article preserves both conventions as they appear in their source references.} A survey of representative definitions for each is provided in Table~\ref{tab:Awareness}.

\begin{table}[!htbp]
\centering
\begin{tabularx}{\textwidth}{l X}
\toprule
\multicolumn{2}{c}{\textbf{Mutual Awareness}} \\
\midrule
\textbf{Source} & \textbf{Definition} \\
\midrule
\cite{MutualJeanPaley} & The extent to which team members are informed of other team members' behaviors \\
\midrule
\cite{MutualSchuster} & The level of understanding participants have of each other's actions and contexts, which improves efficiency and performance in a simulator. \\
\midrule
\cite{MutualRognin} & Awareness about each other's activities and about each other's awareness. \\
\midrule
\cite{MA-Notes} & The perception and understanding of member A on the activity, including intention, status, possible results and influence, of member B. \\
\midrule
\cite{VillaAwareness2008} & If member A could observe member B's activities while member B could not observe A, then A has a watching awareness. If member A knows member B is observing him/her, but A himself/herself could not watch B, then A has a watched awareness. \\
\midrule
\multicolumn{2}{c}{\textbf{Shared Situational Awareness}} \\
\midrule
\textbf{Source} & \textbf{Definition} \\
\midrule
\multirow{2}{*}{\cite{Endsley_Intra_Inter_TSA}} & The degree to which team members possess the same SA on shared SA requirements. \\
\cmidrule(l){2-2}
 & Shared common understanding among the team members of relevant SA elements. \\
\midrule
\cite{salasrole} & A continuously updated and integrated mental representation of a mission built from shared information among team members. \\
\midrule
\multicolumn{2}{c}{\textbf{Team Situational Awareness}} \\
\midrule
\textbf{Source} & \textbf{Definition} \\
\midrule
\cite{Endsley_Intra_Inter_TSA} & The degree to which every team member possesses the SA required for his or her responsibilities. \\
\midrule
\cite{Shu_Inference_TeamSA} & Two or more individuals share the common environment, up-to-the-moment understanding of the environment, and another person's interaction with the cooperative task. \\
\midrule
\cite{SalasTSA1995} & Team SA involves two critical but poorly understood abstractions: individual SA and team processes (\ie teamwork behaviors and cognitive processes that facilitate team performance). \\
\bottomrule
\end{tabularx}
\caption{Definitions of Mutual Awareness, Shared Situational Awareness, and Team Situational Awareness in collective contexts.}
\label{tab:Awareness}
\end{table}

\subsection{Mutual Awareness}

Mutual Awareness (MA) is described as the mutual understanding of each other's intentions, beliefs and activities~\cite{TSA_Survey_She}. Schmidt~\cite{MA-Notes} characterizes it more precisely as the perception and understanding that member A has of the activity, intention, status, possible results, and influence of member B. This definition highlights that MA is inherently directional and may be asymmetric: as noted by Villa et al.~\cite{VillaAwareness2008}, if A can observe B but B cannot observe A, then A holds a \emph{watching awareness}, whereas if A knows B is observing them but cannot reciprocate, A holds a \emph{watched awareness}. Some theories propose that MA is essential for any cooperative activity~\cite{Shu_Inference_TeamSA}, and several authors reduce MA to individual activity and mutual beliefs~\cite{Grosz_Plans_For_Discourse,We-Intentions}, which influence the thoughts and actions of team members. Similarly, Bratman~\cite{SCA_Bratman_92} states that mutual responsiveness, defined as responsiveness to the intentions and actions of others, which emerges when participants are committed to the same goal, is essential for achieving a shared cooperative activity. This matches the idea proposed in~\cite{Grosz_Plans_For_Discourse} regarding mutual beliefs about the intentions of other individuals being considered in the definition of a shared plan. Complementary to individual SA, mutual awareness is also mentioned in~\cite{TSA_Survey_She} as a component of TSA, where authors describe MA as the understanding of each other's activities, beliefs and intentions.

\subsection{Shared Situational Awareness}

Shared Situational Awareness (SSA) refers to the overlap between the SA requirements of different team members, or the degree to which team members possess the same SA across shared requirements~\cite{Endsley_Intra_Inter_TSA}. Shared SA is also described as the overlap in situation awareness that is necessary among team members~\cite{Endsley_Intra_Inter_TSA}. Complementing this view, Salas et al.~\cite{salasrole} describe Shared SA as a continuously updated and integrated mental representation of a mission built from shared information among team members, emphasizing its dynamic and constructive character. Endsley also suggests that teams can develop internal mechanisms to facilitate the achievement of shared SA. Furthermore, Endsley and Jones state that the information to be shared across teams should be defined by examining the areas where SA requirements of each team overlap, to avoid excessive communication overload. This shared information is specifically characterized as critical~\cite{SSASeppanen}.

\subsection{Team Situational Awareness}

Team Situational Awareness (TSA) has been the subject of varied interpretations, reflecting differing assumptions about how awareness is formed and maintained in teams. Some definitions of TSA describe it primarily as the combination or sharing of individual team members' situational awareness (SA). From this perspective, TSA corresponds to the overlap of individual SA knowledge states~\cite{TSA_NathanJ}. A common definition captures TSA as the degree to which every team member possesses the SA necessary for fulfilling their responsibilities~\cite{ENDSLEY2000301} including not only the environment but also each individual's interaction with the cooperative task~\cite{Shu_Inference_TeamSA}. Salas et al.~\cite{SalasTSA1995} further identify two critical abstractions underlying TSA: individual SA and team processes, where the latter encompasses the teamwork behaviors and cognitive processes that facilitate team performance and that cannot be reduced to the sum of individual awareness states. In contrast, Salas et al. describe TSA as a shared understanding of the situation among team members at a given moment~\cite{Salas_SA}. Similarly, another viewpoint characterizes it as a common perspective among two or more individuals regarding environmental events, their implications, and likely developments~\cite{TSA_Survey_She}. Artman and Garbin~\cite{artman1998situation} expand on this, defining TSA as a collaboratively constructed model that is partly shared and partly distributed, enabling anticipation of future states. Finally, Stanton et al. describe TSA as knowledge of a dynamic organizational structure and environment, actively formed and sustained through interactions with both the external world and other team members~\cite{Stanton03042017}.

\section{Taxonomy of Collective Awareness} \label{section:CATaxonomy}

To provide a structured characterization of the awareness types discussed in the previous section, this section presents a taxonomy organized along two key axes: the object of awareness and awareness distribution. Before examining how each awareness type is positioned along these axes, we introduce the illustrative example that will be used throughout this section to ground the abstract classifications in a concrete setting.

The example concerns a heterogeneous robotic team cooperating in a Search and Rescue (SaR) operation. The team is composed of aerial robots (UAVs) equipped with cameras and image analysis capabilities, and ground robots (UGVs) equipped with manipulation capabilities. The mission unfolds across four interdependent phases: deployment, in which robots are dispatched to the affected area; reconnaissance, in which robots survey the environment, build maps, and search for survivors; rescuing, in which heterogeneous robots collaborate to remove debris and assist survivors; and monitoring, in which robots track survivors' conditions and environmental hazards. These phases are not strictly sequential and often overlap, requiring the team to maintain awareness at multiple levels simultaneously under conditions of unreliable communication, dynamic environmental changes, and role-based task specialization. Concretely, during reconnaissance, UGVs must navigate debris-filled corridors without interfering with each other; UAVs detect and localize survivors and broadcast their location and assessed status to the whole team; and UAVs identify safe landing zones using aerodynamic models specific to aerial platforms and share these assessments with other UAVs for mission replanning. This scenario provides a concrete reference for illustrating how each awareness type is positioned along the taxonomy axes. The following figures illustrate how each awareness type manifests concretely within this setting.

\begin{figure}[tbh]
    \centering
    \includegraphics[width=\linewidth]{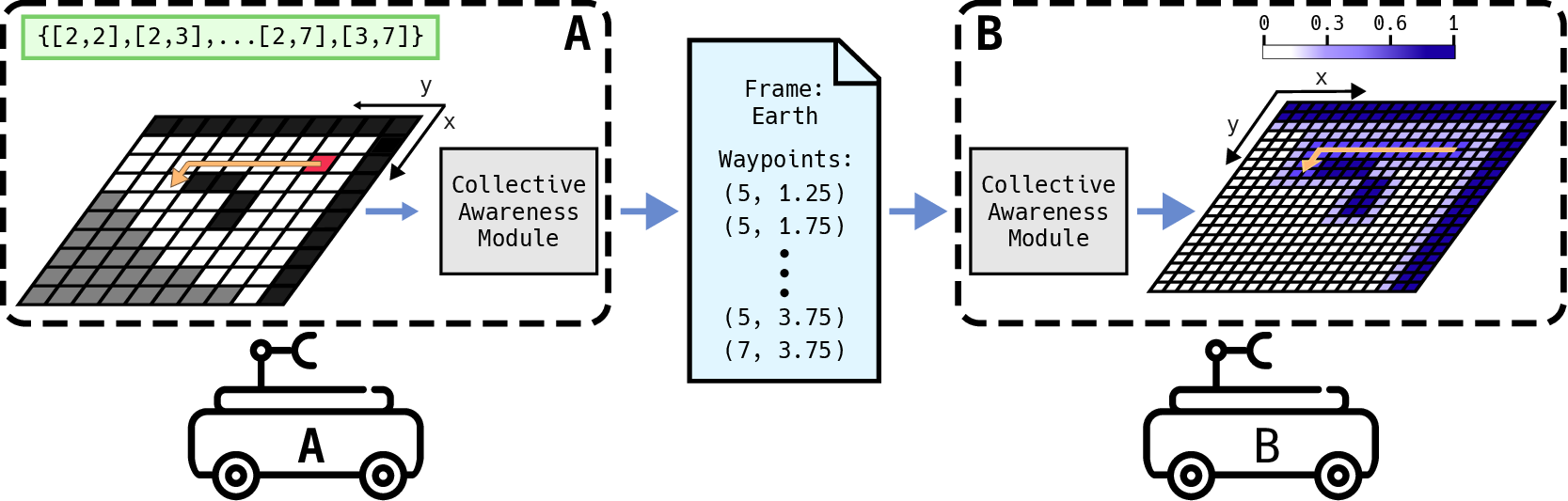}
    \vspace{4mm}
    \caption{UGV A plans a trajectory in its local occupancy grid and sends the cell coordinates to its CA module, which transforms them into coordinates in a common reference frame. The CA module transmits this shared representation to UGV B, whose CA module transforms the Cartesian coordinates into cells in B's local cost map and updates them accordingly. This coordinate transformation chain enables mutual awareness of planned paths despite heterogeneous internal representations.}
    \label{fig:MAEx}
\end{figure}

Figure~\ref{fig:MAEx} illustrates mutual awareness during the reconnaissance phase. Two UGVs with heterogeneous internal representations must coordinate their navigation through debris-filled corridors. UGV A plans a trajectory in its local occupancy grid and communicates it through the Collective Awareness (CA) module; UGV B receives the trajectory and integrates it into its own costmap representation. Each robot forms an individualized understanding of the peer's planned path, adapted to its own internal coordinate system and navigation needs. The object of awareness is the planned trajectory of a peer agent, and each robot maintains its own localized interpretation of that trajectory rather than converging on a common representation, which is the defining characteristic of mutual awareness.

\begin{figure}[tbh]
    \centering
    \includegraphics[width=\linewidth]{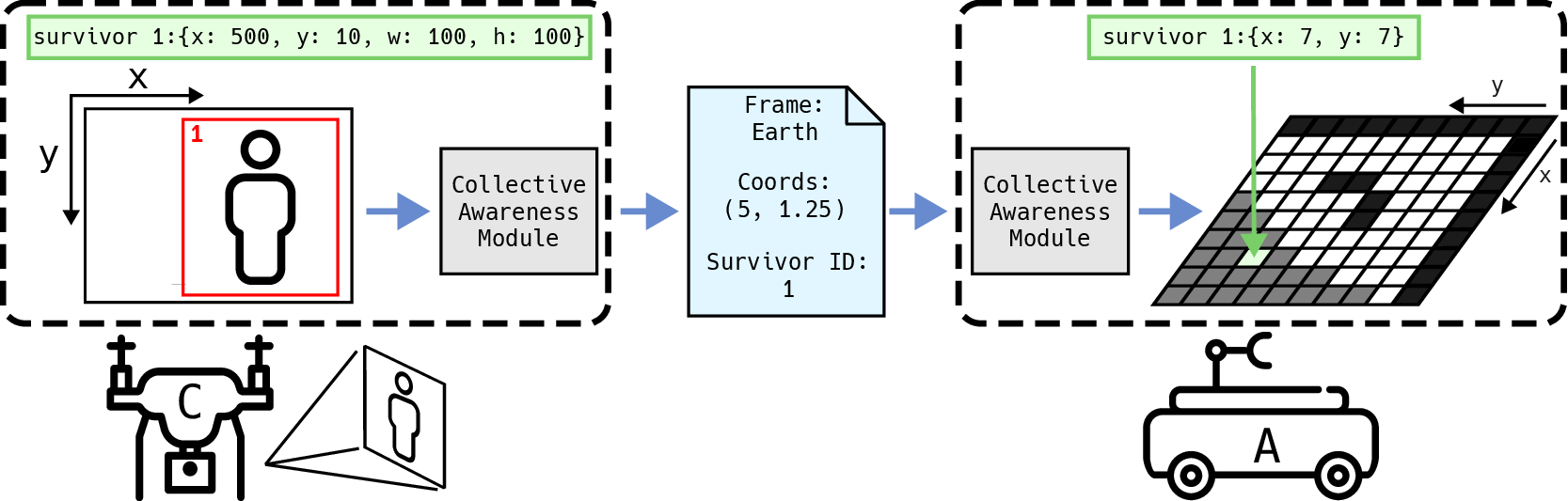}
    \vspace{4mm}
    \caption{A UAV detects a survivor in an image and obtains bounding box coordinates in the image frame. The UAV's CA module transforms these coordinates into real coordinates in a common reference frame and transmits them to a UGV. The UGV's CA module transforms the received coordinates into a cell in its local occupancy grid, enabling the UGV to integrate the survivor location into its spatial representation. This transformation chain allows heterogeneous agents to share situational awareness despite different sensing modalities and internal representations.}
    \label{fig:SSAEx}
\end{figure}

Figure~\ref{fig:SSAEx} illustrates shared situational awareness during the same reconnaissance phase. A UAV detects a survivor and obtains their position in the image frame; the CA module transforms these coordinates into a common reference frame and transmits them to the rest of the team. Each receiving agent converts the shared coordinates into its own internal representation, but the semantic content is identical across agents: the identity, location, and assessed status of the survivor. The object of awareness is a mission-critical environmental element that all agents must integrate and agree upon uniformly, regardless of their individual internal representations, which is the defining characteristic of shared situational awareness.

\begin{figure}[tbh]
    \centering
    \includegraphics[width=\linewidth]{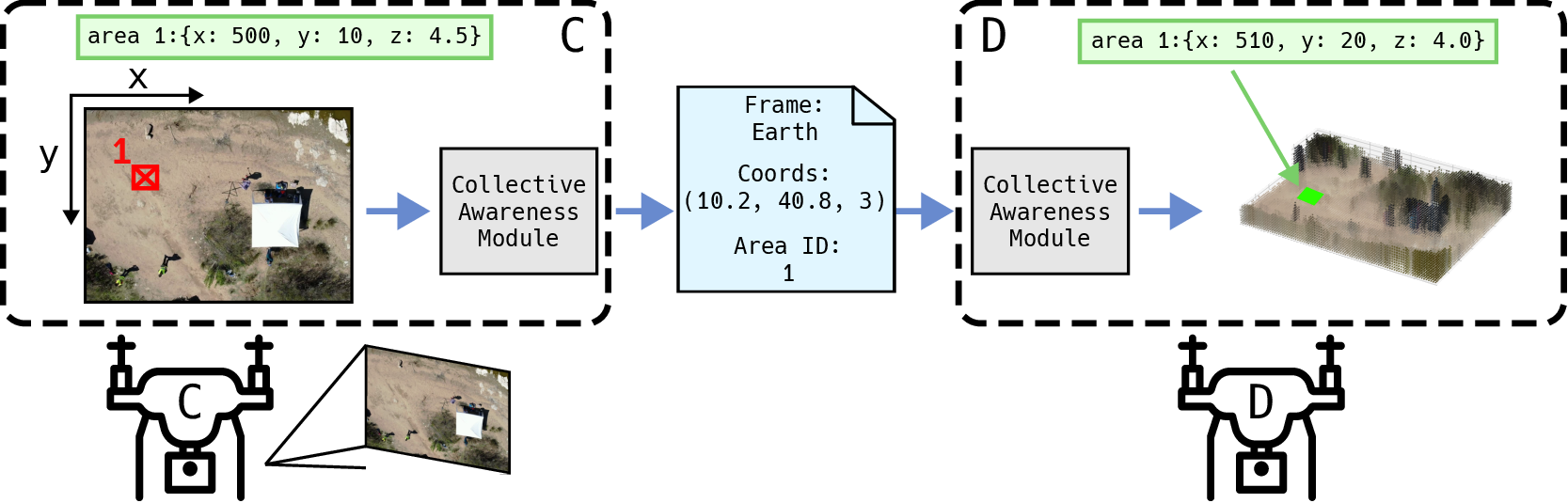}
    \vspace{4mm}
    \caption{UAV C detects a safe landing area in an image captured by its camera and obtains bounding box coordinates in the image frame. C's collective awareness module transforms these coordinates into real Cartesian coordinates in a common reference frame and transmits them to another UAV. The receiving UAV's CA module transforms the shared coordinates into voxels in its local voxel map, enabling it to integrate the landing zone location into its spatial representation. This shared understanding of a critical environmental feature supports coordinated team operations and mission planning.}
    \label{fig:TSAEx}
\end{figure}

Figure~\ref{fig:TSAEx} illustrates team situational awareness during the mission replanning phase. A UAV identifies a safe landing zone using its camera and aerodynamic models, transforms the location into a common reference frame, and shares it with other UAVs. Ground robots do not require this information, as landing zone assessment lies outside their role-specific awareness requirements. The landing zone information, combined with survivor locations and navigation paths shared across the full team, forms the complete team situational awareness of the mission: a layered structure of elements shared by all agents and elements specific to each platform type, which is the defining characteristic of team situational awareness.

The first axis of the taxonomy, the object of awareness, identifies the elements that agents must know about within the collaborative system. The second axis, awareness distribution, describes how awareness is spread across the team: specifically, whether agents converge on identical conclusions from a given set of information or whether their understanding diverges according to their individual roles and perspectives. Together, these two axes allow the awareness concepts reviewed in Section~\ref{sectionRevAwareness} to be analyzed not merely as labels but as markers of distinct requirements for information exchange and coordination. The following subsections examine how Mutual Awareness, Shared Situational Awareness, and Team Situational Awareness are positioned along each axis. The results of this classification are summarized in Table~\ref{tab:AwCat}.

\begin{table}[tbh]
    \centering
    \begin{tabularx}{\textwidth}{l l X}
         \toprule
         \textbf{Type of Awareness} &  \textbf{Distribution} & \textbf{Object of Awareness} \\
         \midrule
         Mutual Awareness &  Distributed & Team member's intentions, beliefs, goals, and plans. \\
         \midrule
         Shared Situational Awareness  & Shared & Essential elements for coordinated task execution. The intersection of each individual agent's awareness requirements. \\
         \midrule
         Team Situational Awareness  & Mixed & Individualized awareness each agent must maintain to effectively perform their own functions. Overlaps with agents performing the same functions. \\
         \bottomrule
    \end{tabularx}
    \caption{Categorization of each term related to awareness in multi-agent systems according to the distribution and object of the awareness.}
    \label{tab:AwCat}
\end{table}

\subsection{Object of Awareness}

Beyond the definitions discussed in Section~\ref{sectionRevAwareness}, it is important to examine what agents or team members can be aware of within a collaborative system. Identifying the elements in the environment or within the team that constitute the content of awareness is essential because the breadth and nature of these objects evolve alongside the system's increasing awareness capabilities. As the awareness of the team becomes more sophisticated, agents transition from being aware of isolated features, such as the trajectory or state of neighboring agents, to developing an integrated understanding of the entire operational context, including tasks, roles, goals, and dynamic environmental conditions.

\begin{figure}[tbh]
    \centering
    \includegraphics[width=0.6\linewidth]{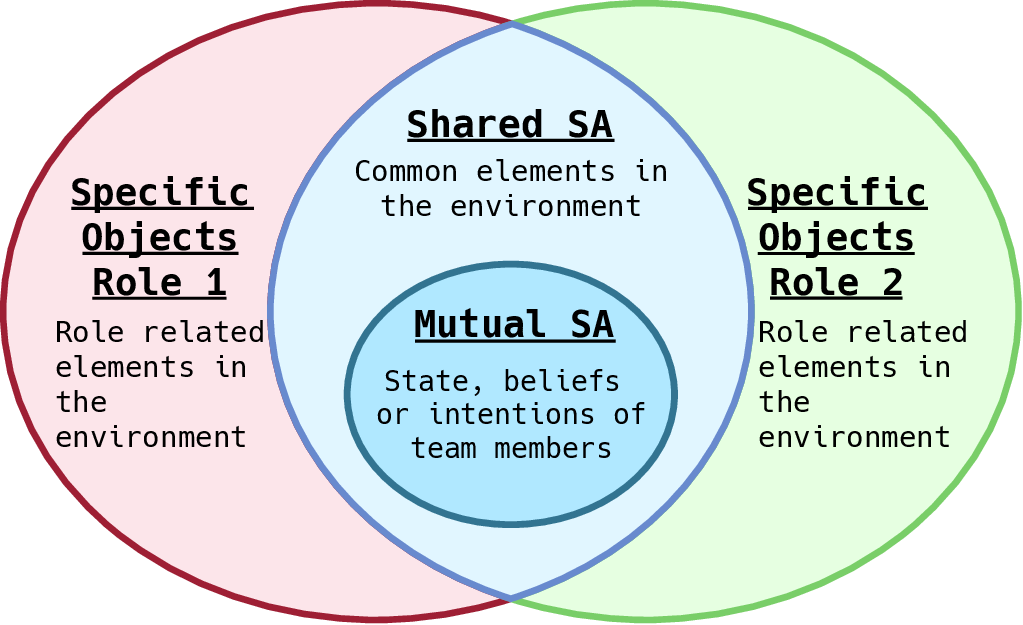}
    \vspace{4mm}
    \caption{Venn diagram of the objects of mutual awareness, shared situational awareness, and role-specific elements. All of them combined are objects of team situational awareness.}
    \label{fig:AwTypesAbs}
\end{figure}

A parallel effort to enumerate objects of awareness in human–robot teams, organized along a complementary directional axis, was independently developed by~\cite{DrurySMC2003}.
This subsection categorizes the three awareness types based on the scope and nature of the elements they concern. Figure~\ref{fig:AwTypesAbs} illustrates the hierarchical relationship among their respective objects of awareness.

When examining the concept of mutual awareness, its primary objects are the other members of the team and their internal states, such as intentions, beliefs, goals, and plans. Mutual awareness emphasizes the reciprocal understanding that agents have of one another, which is fundamental to effective teamwork. This form of awareness is not only about knowing that other agents exist or recognizing their presence in the environment, but also involves awareness of what other agents are trying to achieve, how they interpret the current situation, and how they are likely to act based on that interpretation. Leveraging this awareness of other team members, agents can better anticipate the behavior of their teammates, adapt their own actions accordingly, and proactively coordinate in pursuit of shared objectives. In the SaR example, the object of mutual awareness is the internal state of a peer agent, specifically its intended trajectory through the operational environment (Figure~\ref{fig:MAEx}).

The concept of shared situational awareness refers to the common understanding among team members of those elements that are essential for coordinated task execution, specifically the elements that fall within the intersection of what each individual agent is required to be aware of to fulfill their role. This means that Shared SA involves not only an understanding of each agent's internal states, such as their goals, intentions, and actions, but also a collectively held awareness of critical external factors in the environment. As Endsley and Jones explain in~\cite{Endsley_Intra_Inter_TSA}, effective collaboration requires that team members be aware of one another, making mutual awareness a subset of SSA. Shared situational awareness goes beyond mutual awareness by also encompassing elements external to the team, such as mission objectives, environmental constraints, or dynamic threats, that all agents must align on to operate cohesively, making decisions based on a shared and synchronized understanding of the operational context. In the SaR example, the object of shared situational awareness is the location and assessed status of survivors detected during reconnaissance: mission-critical elements that all agents must integrate uniformly regardless of their individual spatial representations (Figure~\ref{fig:SSAEx}).

Expanding to an even broader domain, Team Situation Awareness builds upon the foundations of both mutual and shared situational awareness by incorporating the structure and functional diversity within a team. As emphasized by Endsley and Jones in \cite{Endsley_Intra_Inter_TSA}, one of the defining characteristics of a team is the presence of distinct roles, each associated with specific responsibilities and subgoals that contribute to the achievement of a common team objective. These role-based subgoals naturally give rise to varying information needs among agents, meaning that the situation awareness required by each team member is shaped not only by what is commonly shared across the team, but also by the unique demands of their individual roles. Consequently, team situational awareness encompasses both the shared awareness necessary for coordinated collaboration, captured by SSA, and the individualized awareness each agent must maintain to effectively perform their own functions. This layered structure of awareness ensures that while all agents are aligned on critical common elements, they also possess specialized knowledge tailored to their specific contributions, allowing the team as a whole to function as a coherent and adaptive unit. In the SaR example, TSA integrates the shared mission picture with role-specific elements: UAVs must additionally maintain awareness of wind conditions and potential landing zones assessed through aerodynamic models, while UGVs must track debris properties and terrain traversability specific to ground navigation. These role-specific objects of awareness complement the shared elements to form the complete awareness landscape required for effective mission execution (Figure~\ref{fig:TSAEx}).

\begin{figure}[tbh]
     \centering
     \includegraphics[width=.6\linewidth]{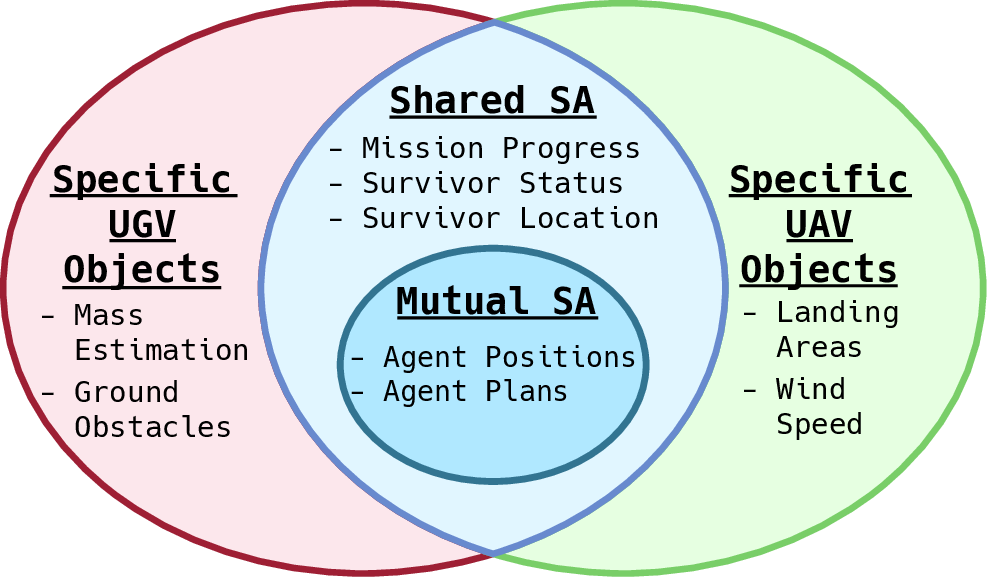}
     \caption{Venn diagram of the objects of each kind of awareness for a system made up of ground and aerial robots operating in a search and rescue scenario. Specific objects of awareness for each role, combined with mutual awareness and shared situational awareness objects, form the whole set of objects of team situational awareness.}
     \label{fig:AwTypes}
\end{figure}

Figure~\ref{fig:AwTypes} illustrates the complete landscape for the SaR scenario, showing how the objects of each awareness type combine to cover the full set of information the team must maintain.

\subsection{Awareness Distribution}

A second dimension along which the three awareness types can be differentiated concerns how awareness is spread across agents in the team. At one end of the spectrum, awareness may be shared: all agents reach the same conclusions and maintain the same understanding of a given element, eliminating the need for ongoing coordination about those elements. At the other end, awareness may be distributed: agents form individualized understandings of the same element, shaped by their own roles, perspectives, and available information~\cite{Endsley_Intra_Inter_TSA}~\cite{Stanton_DSA}.

We define awareness distribution for a multi-agent system as the degree to which awareness of a given element is uniform across agents in the system. This distribution can manifest in different forms depending on the nature of the awareness being studied and the coordination demands of the task at hand.

Mutual awareness belongs to the distributed end of the spectrum, as it simply requires agents to be aware of each other's presence, actions, or intentions without enforcing any alignment in their awareness of specific elements in the environment. There is no requirement for agents to share the same understanding or inferences, allowing each to maintain individualized awareness shaped by their roles and perspectives. In the SaR example, this distributed character is evident in how two UGVs track each other's planned trajectories during reconnaissance. One robot may represent its environment as a 2D occupancy grid while the other uses a costmap; each forms its own localized understanding of the peer's intended path without requiring identical internal representations (Figure~\ref{fig:MAEx}).

Shared situational awareness, in contrast, is placed at the shared end of the spectrum, as it demands that all agents maintain the same awareness regarding particular critical elements, supporting seamless coordination through identical interpretations of key aspects of the environment. In the SaR example, when a UAV detects and localizes a survivor, it broadcasts the location and assessed status to the entire team. Although each robot integrates this data into its own internal representation, the logical content, that is, the identity, location, and status of the survivor, is identical across all agents, reflecting the shared character of this awareness type (Figure~\ref{fig:SSAEx}).

Team situational awareness lies between these two extremes: it requires shared awareness of certain task-relevant elements necessary for coordinated action, while allowing individual agents to maintain divergent or role-specific awareness of other elements not critical to joint decision-making. This partial alignment enables flexibility and task specialization while still supporting cohesive team behavior. In the SaR example, all robots share awareness of survivor locations and mission progress, while UAVs additionally maintain their own role-specific assessments of airspace conditions and safe landing zones. This combination of shared and individualized awareness allows the team to coordinate effectively while each platform contributes specialized knowledge that the other cannot produce (Figure~\ref{fig:TSAEx}).

\subsection{Synthesized Definitions}

Drawing on the two axes established above, we state a precise definition for each awareness type that captures both its object and its distribution.

\begin{definition}[Mutual Awareness]
The awareness each agent holds of its teammates' internal states, including their intentions, beliefs, goals, and plans, where each agent forms an individualized understanding shaped by its own perspective and role.
\end{definition}

In the SaR scenario, mutual awareness governs how UGVs track each other's planned trajectories during reconnaissance. When UGV A communicates its path, UGV B integrates that information into its own costmap, forming an understanding of A's intentions adapted to its own internal architecture (Figure~\ref{fig:MAEx}). The object of awareness is the peer's planned path; the distribution is individualized, as B's representation need not match A's internal encoding, only capture the same spatial intent in B's own terms.

\begin{definition}[Shared Situational Awareness]
The awareness all agents in a team must maintain of the environmental and mission elements that fall within every agent's requirements, specifically their intersection, such that all agents reach an aligned interpretation of the information essential for coordinated action.
\end{definition}

When a UAV detects a survivor and broadcasts their location and assessed status, every agent must integrate the same conclusion regardless of its internal representation: a survivor is present at a specific location with a given urgency (Figure~\ref{fig:SSAEx}). The object of awareness is the survivor's location and status; the distribution is shared, as no agent may hold a divergent interpretation of this element without risking conflicting rescue priorities.

\begin{definition}[Team Situational Awareness]
The complete awareness a robotic team must maintain, encompassing both the shared understanding of mission-critical elements common to all agents (SSA) and elements of the environment that are relevant to the mission yet specific to certain roles, such as platform-dependent terrain or airspace conditions, for which divergent, role-shaped understanding is appropriate.
\end{definition}

The full awareness landscape of the SaR mission integrates the shared elements above with role-specific knowledge that is mission-relevant but not required by every platform. UAVs must assess wind conditions and identify safe landing zones; UGVs must track debris properties and terrain traversability (Figure~\ref{fig:TSAEx}). Neither is captured by MA or SSA alone, yet both are necessary for effective mission execution. TSA encompasses all three layers: peer-state awareness, shared mission-critical awareness, and role-specific environmental knowledge.

\section{Conclusions} \label{section:Conclusions}

This article has addressed two fundamental questions about collective awareness in multi-robot systems, contributing a conceptual framework that clarifies the theory-practice gap surrounding awareness terminology in multi-robot coordination.

\begin{enumerate}
  \item \textit{What are the distinct requirements of the different awareness types, and how do they relate to each other?} \\ Our structured taxonomy categorizes collective awareness along two dimensions, awareness distribution and object of awareness, establishing that mutual awareness, shared situational awareness, and team situational awareness form a layered hierarchy with distinct requirements. MA requires that agents maintain awareness of each other's beliefs, plans, and states to enable basic coordination. SSA builds upon MA by demanding shared semantic understanding of mission-critical environmental elements, ensuring aligned interpretation across the team. TSA integrates both shared awareness and role-specific knowledge, enabling specialized coordination while maintaining collective coherence. The Search and Rescue case study illustrates how mission requirements decompose into these specific awareness specifications, showing that each phase demands different combinations of awareness types. \\

  \item \textit{What precise definitions can be established for each awareness type that characterize their distinct requirements and roles in multi-robot coordination?} \\ The taxonomy yields the three definitions stated in Section~\ref{section:CATaxonomy}, each grounded in both axes. Together they establish that MA, SSA, and TSA form a layered structure with distinct distribution and content requirements, rather than interchangeable labels for the same phenomenon. \\

\end{enumerate}

This taxonomy provides a conceptual foundation for more principled design and meaningful comparison of multi-robot coordination approaches. Despite the conceptual clarity this taxonomy provides, several open challenges remain. Future work should formalize the requirements identified by the taxonomy as verifiable system specifications, develop architectural components that instantiate these requirements, and empirically evaluate their effect on coordination performance in deployed systems.

\section*{Declarations}

\begin{itemize}
    \item Acknowledgements: This work has been supported by the project "CORESENSE: A Hybrid Cognitive Architecture for Deep Understanding" Ref. No. 101070254 , funded by European Union under the Horizon Europe Program, and the project INSERTION Grant ref. ID2021-127648OBC32 funded by MICIU/AEI/ 10.13039/501100011033 and by FEDER, EU and RETMUR Grant ref:  RED2022-134707-T funded by MICIU/AEI /10.13039/501100011033. The work of the first author is supported by the Community of Madrid under its Program for Predoctoral Researcher with Ref. PIPF-2023/TEC-31167 (BOCM - 3454/2024) \\

    \item Funding: This work has been supported by the project "CORESENSE: A Hybrid Cognitive Architecture for Deep Understanding"   Ref. No. 101070254 , funded by European Union under the Horizon Europe Program, and the project INSERTION Grant ref. ID2021-127648OBC32 funded by MICIU/AEI/ 10.13039/501100011033 and by FEDER, EU and RETMUR Grant ref:  RED2022-134707-T funded by MICIU/AEI /10.13039/501100011033. The work of the first author is supported by the Community of Madrid under its Program for Predoctoral Researcher with Ref. PIPF-2023/TEC-31167 (BOCM - 3454/2024)\\

    \item Competing interests: The authors have no competing interests to declare that are relevant to the content of this article. \\

    \item Ethics approval and consent to participate: Not applicable\\
    \item Consent for publication: All authors explicitly consent to submit and publish the work\\
    \item Data availability: Not applicable\\
    \item Materials availability: Not applicable\\
    \item Code availability: Not applicable\\
    \item Author contribution: All authors contributed to the study conception and design. Material preparation, literature review and analysis were performed by Guillermo GP-Lenza. The first draft of the manuscript was written by Guillermo GP-Lenza and all authors commented on previous versions of the manuscript. All authors read and approved the final manuscript.

\end{itemize}

\section*{Compliance with Ethical Standards}

The authors declare that there are not conflicts of interests related with the content of the article, and no experiments have been conducted on human participants or animals.

\bibliography{sn-bibliography}

\end{document}